\documentclass[10pt,twocolumn,letterpaper]{article}

\usepackage{iccv}
\usepackage[accsupp]{axessibility}
\usepackage{times}
\usepackage{epsfig}
\usepackage{graphicx}
\usepackage{amsmath}
\usepackage{amssymb}
\usepackage{mathbbol}
\usepackage{algorithmic}
\usepackage[ruled,vlined]{algorithm2e}
\usepackage{multirow}
\usepackage{multicol} 
\usepackage{array}
\usepackage{makecell}
\usepackage[dvipsnames]{xcolor}
\usepackage{booktabs}
\usepackage{arydshln}

\definecolor{improvement}{rgb}{0.4, 0.55, 0.3}

\usepackage[pagebackref=true,breaklinks=true,letterpaper=true,colorlinks,bookmarks=false]{hyperref}

\iccvfinalcopy % *** Uncomment this line for the final submission
\def\iccvPaperID{2614} % *** Enter the ICCV Paper ID here

\ificcvfinal\fi

\begin{document}

	% signs
	\newcommand{\fracpartial}[2]{\frac{\partial #1}{\partial  #2}}
	\newcommand{\norm}[1]{\left\lVert#1\right\rVert}
	\newcommand{\innerproduct}[2]{\left\langle#1, #2\right\rangle}
	\newcommand{\fan}[1]{\Vert #1 \Vert}
	\newcommand{\qileft}{[\kern-0.15em[}
	\newcommand{\qiLeft}{\left[\kern-0.4em\left[}
	\newcommand{\qiright}{]\kern-0.15em]}
	\newcommand{\qiRight}{\right]\kern-0.4em\right]}
	\newcommand{\sign}{{\mbox{sign}}}
	\newcommand{\diag}{{\mbox{diag}}}
	\newcommand{\armin}{{\mbox{argmin}}}
	\newcommand{\rank}{{\mbox{rank}}}
	\renewcommand{\vec}{{\mbox{vec}}}
	\newcommand{\st}{{\mbox{s.t.}}}
	\newcommand{\<}{\left\langle}
	\renewcommand{\>}{\right\rangle}
	\newcommand{\lbar}{\left\|}
	\newcommand{\rbar}{\right\|}
	\renewcommand{\Roman}[1]{\uppercase\expandafter{\romannumeral#1}}
	\newcommand{\red}[1]{{\color{red}{#1}}}
	\newcommand{\blue}[1]{{\color{blue}{#1}}}

	% for vectors
	\renewcommand{\a}{{\bm{a}}}
	\renewcommand{\b}{{\bm{b}}}
	\renewcommand{\d}{{\bm{d}}}
	\newcommand{\e}{{\bm{e}}}
	\newcommand{\f}{{\bm{f}}}
	\newcommand{\g}{{\bm{g}}}
	\renewcommand{\o}{{\bm{o}}}
	\newcommand{\p}{{\bm{p}}}
	\newcommand{\q}{{\bm{q}}}
	\renewcommand{\r}{{\bm{r}}}
	\newcommand{\s}{{\bm{s}}}
	\renewcommand{\t}{{\bm{t}}}
	\renewcommand{\u}{{\bm{u}}}
	\renewcommand{\v}{{\bm{v}}}
	\newcommand{\w}{{\bm{w}}}
	\newcommand{\x}{{\bm{x}}}
	\newcommand{\y}{{\bm{y}}}
	\newcommand{\z}{{\bm{z}}}
	\newcommand{\balpha}{{\bm{\alpha}}}
	\newcommand{\bbeta}{{\bm{\beta}}}
	\newcommand{\bmu}{{\bm{\mu}}}
	\newcommand{\bsigma}{{\bm{\sigma}}}
	\newcommand{\blambda}{{\bm{\lambda}}}
	\newcommand{\btheta}{{\bm{\theta}}}
	\newcommand{\bgamma}{{\bm{\gamma}}}
	\newcommand{\bxi}{{\bm{\xi}}}
	\newcommand{\bphi}{{\bm{\phi}}}
	
	% for matrixes bold
	\newcommand{\ba}{{\bm{A}}}
	\newcommand{\bb}{{\bm{B}}}
	\newcommand{\bc}{{\bm{C}}}
	\newcommand{\bd}{{\bm{D}}}
	\newcommand{\be}{{\bm{E}}}
	\newcommand{\bg}{{\bm{G}}}
	\newcommand{\bi}{{\bm{I}}}
	\newcommand{\bj}{{\bm{J}}}
	\newcommand{\bl}{{\bm{L}}}
	\newcommand{\bo}{{\bm{O}}}
	\newcommand{\bp}{{\bm{P}}}
	\newcommand{\bq}{{\bm{Q}}}
	\newcommand{\bs}{{\bm{S}}}
	\newcommand{\bu}{{\bm{U}}}
	\newcommand{\bv}{{\bm{V}}}
	\newcommand{\bw}{{\bm{W}}}
	\newcommand{\bx}{{\bm{X}}}
	\newcommand{\by}{{\bm{Y}}}
	\newcommand{\bz}{{\bm{Z}}}
	\newcommand{\bTheta}{{\bm{\Theta}}}
	\newcommand{\bSigma}{{\bm{\Sigma}}}
	
	% for matrixes mathcal
	\newcommand{\A}{{\mathcal{A}}}
	\newcommand{\B}{\mathcal{B}}
	\newcommand{\C}{\mathcal{C}}
	\newcommand{\D}{\mathcal{D}}
	\newcommand{\F}{\mathcal{F}}
	\renewcommand{\H}{\mathcal{H}}
	\newcommand{\I}{\mathcal{I}}
	\renewcommand{\L}{\mathcal{L}}
	\newcommand{\N}{\mathcal{N}}
	\renewcommand{\P}{\mathcal{P}}
	\newcommand{\X}{\mathcal{X}}
	\newcommand{\Y}{\mathcal{Y}}
	\newcommand{\W}{\mathcal{W}}

%%%%%%%%% TITLE
\title{Weakly Supervised Contrastive Learning \vspace{-8pt}}

% \author{Xiu Su$^{1}$, Shan You$^{2,3}$\thanks{Corresponding authors.}, ~Fei Wang$^{2}$, Chen Qian$^{2}$, Changshui Zhang$^{3}$, Chang Xu$^{1*}$\\
% $^1$School of Computer Science, Faculty of Engineering, The University of Sydney, Australia\\
% $^2$SenseTime Research\\
% $^3$Department of Automation, Tsinghua University,\\
% Institute for Artificial Intelligence, Tsinghua University (THUAI), \\
% Beijing National Research Center for Information Science and Technology (BNRist) \\
% {\tt\small xisu5992@uni.sydney.edu.au, \{youshan,wangfei,qianchen\}@sensetime.com}\\
% {\tt\small zcs@mail.tsinghua.edu.cn, c.xu@sydney.edu.au}
% }

\author{
	Mingkai Zheng$^{1*}$\quad Fei Wang$^{2}$\thanks{Equal contributions.}\quad Shan You$^{1,3}$\thanks{Corresponding author.}\quad Chen Qian$^1$ \\ Changshui Zhang$^{3}$\quad Xiaogang Wang$^{1,4}$\quad Chang Xu$^5$ \\
	$^1$SenseTime Research \quad $^2$University of Science and Technology of China \\
	 $^{3}$Department of Automation, Tsinghua University,\\
Institute for Artificial Intelligence, Tsinghua University (THUAI), \\
Beijing National Research Center for Information Science and Technology (BNRist)\\
\quad	 $^4$The Chinese University of Hong Kong\\
	$^5$School of Computer Science, Faculty of Engineering, The University of Sydney\\
	 {\tt\small\{zhengmingkai,youshan,qianchen\}@sensetime.com},~ {\tt\small wangfei91@mail.ustc.edu.cn}\\
 {\tt\small zcs@mail.tsinghua.edu.cn}, {\tt\small xgwang@ee.cuhk.edu.hk}, {\tt\small c.xu@sydney.edu.au}\\ 
 \vspace{-17pt}
}

\maketitle
\ificcvfinal\thispagestyle{empty}\fi
%%%%%%%%% ABSTRACT
\begin{abstract}
\vspace{-5pt}
Unsupervised visual representation learning has gained much attention from the computer vision community because of the recent achievement of contrastive learning. Most of the existing contrastive learning frameworks adopt the instance discrimination as the pretext task, which treating every single instance as a different class. However, such method will inevitably cause class collision problems, which hurts the quality of the learned representation. Motivated by this observation, we introduced a weakly supervised contrastive learning framework (WCL) to tackle this issue. Specifically, our proposed framework is based on two projection heads, one of which will perform the regular instance discrimination task. The other head will use a graph-based method to explore similar samples and generate a weak label, then perform a supervised contrastive learning task based on the weak label to pull the similar images closer. We further introduced a K-Nearest Neighbor based multi-crop strategy to expand the number of positive samples. Extensive experimental results demonstrate WCL improves the quality of self-supervised representations across different datasets. Notably, we get a new state-of-the-art result for semi-supervised learning. With only 1\% and 10\% labeled examples, WCL achieves 65\% and 72\% ImageNet Top-1 Accuracy using ResNet50, which is even higher than SimCLRv2 with ResNet101.
\end{abstract}
%%%%%%%%% BODY TEXT
\section{Introduction}

Modern deep convolutional neural networks demonstrate outstanding performance on various computer vision datasets \cite{imagenet_cvpr09,pascal-voc-2007,coco} and edge devices \cite{you2017learning,SuYZ00Z021,you2020greedynas,su2021bcnet}. However, most successful methods are trained in the supervised fashion; they usually require a large volume of labeled data that is very hard to collect. Meanwhile, the quality of data annotations dramatically affects the performance. Recently, self-supervised learning shows its superiority and achieves promising results for unsupervised and semi-supervised learning in computer vision (\eg \cite{simclr, simclrv2, moco, mocov2, SimSiam, swav, byol,zheng2021ressl}). These methods can learn general-purpose visual representations without labels and have a good performance on linear classification and transferability to different tasks or datasets.  Notably, a big part of the recent self-supervised representation learning framework is based on the idea of contrastive learning. 

\begin{figure}
    \centering
    \label{fig:motivation}
    \includegraphics[width=0.6\linewidth]{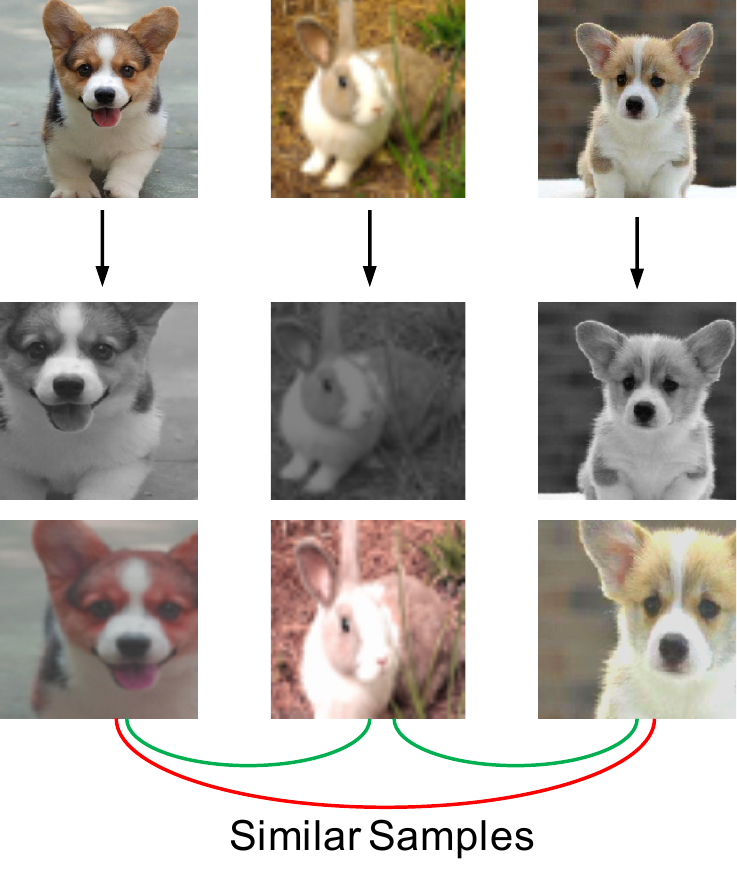}
    \caption{A example of the class collision problem. A typical instance discrimination method will treats the first column and the third column as a negative pair since there are different instance. However, the semantic information of the first column and the third column are very similar, treat them as positive pair should be much more reasonable.}
    \vspace{-15pt}
\end{figure}

A typical contrastive learning based method adopts the noise contrastive estimation (NCE) \cite{cpc} to perform the non-parametric instance discrimination \cite{instance_discrimination} as the pretext task, which encourages the two augmented views of the same image to be pulled closer on the embedding space but pushes apart all the other images. Most of the recent works mainly improve the performance of contrastive learning from the image augmentation for positive samples and the exploration for negative samples. However, instance discrimination based methods will inevitably induce class collision problem, which means even for very similar instances, they still need to be pushed apart, as shown in Figure \ref{fig:motivation}. This instance similarities thus tend to hurt the representation quality \cite{contrastive_theory}. In this way, identifying and even leveraging these similar instances plays a key role in the performance of learned representations.

Surprisingly, the class collision problem seems to attract much lesser attention in contrastive learning. As far as we know, there has been little effort to identify similar samples. AdpCLR \cite{topk} finds the top-K closest samples on the embedding space and treats these samples as their positives. However, in the early stage of training, the model cannot effectively extract the semantic information from the images; therefore, this method needs to use SimCLR \cite{simclr} to pre-train for a period of time, and then switch to AdpCLR to get the best performance. FNCancel \cite{FNCancel} proposed a similar idea but adopts a very different way to find the top-K similar instances; that is, for each sample, it generates a support set that contains different augmented views from the same image, then use mean or max aggregation strategy over the cosine similarity score between the augmented views in support set and finally identify the top-K similar samples. Nevertheless, the optimal support size is 8 in their experiments, requiring 8 additional forwarding passes to generate the embedding vectors. Obviously, these methods have two shortcomings. Firstly, they are both time-consuming. In the second place, the result of top-K closest samples might not be reciprocal, \ie $\mathbf{x}_{i}$ is the K closest sample of $\mathbf{x}_{j}$, but $\mathbf{x}_{j}$ might not be the K closest sample of $\mathbf{x}_{i}$. In this case, $\mathbf{x}_{j}$ will treat $\mathbf{x}_{i}$ as a positive sample, but $\mathbf{x}_{i}$ will treat $\mathbf{x}_{j}$ as a negative sample, which will result in some conflicts.

In this paper, we regard the instance similarities as intrinsically weak supervision in representation learning and propose a weakly supervised contrastive learning framework (WCL) to address the class collision issue accordingly. In WCL, similar instances are assumed to share the same weak label comparing to other instances, and instances with the same weak label are expected to be aggregated. To determine the weak label, we model each batch of instances as a nearest neighbor graph; weak labels are thus determined and reciprocal for each connected component of the graph. Besides, we can further expand the graph by a KNN-based multi-crop strategy to propagate weak labels, such that we can have more positives for each weak label. In this way, similar instances with the same weak label can be pulled closer via the supervised contrastive learning \cite{SupervisedCL} task. Nevertheless, since the mined instance similarities might be noisy and not completely reliable, in practice, we adopt a two-head framework, one of which handles this weakly supervised task while the other is to perform the regular instance discrimination task. Extensive experiments demonstrate the effectiveness of our proposed method across different settings and various datasets.

Our contribution can be summarized as follows:
\begin{itemize}
 \vspace*{-0.3em}
 \setlength\itemsep{-0.1em}
 \item We proposed a two-head based framework to address the class collision problem, with one head focusing on the instance discrimination and the other head for attracting similar samples.
 
  \item We proposed a simple graph based and parameter-free method to find similar samples adaptively. 
 
 \item We introduced a K-Nearest Neighbor based multi-crops strategy that can provide much more diverse information than the standard multi-crops strategy.
 
 \item The experimental result shows WCL establishes a new state-of-the-art performance for contrastive learning based methods. With only 1\% and 10\% labeled samples, WCL achieves 65\% and 72\% Top-1 accuracy on ImageNet using ResNet50.  Notably, this result is even higher than SimCLRv2 with ResNet101.
\end{itemize}

\begin{figure*}[t]
    \centering
    \includegraphics[width=0.8\linewidth]{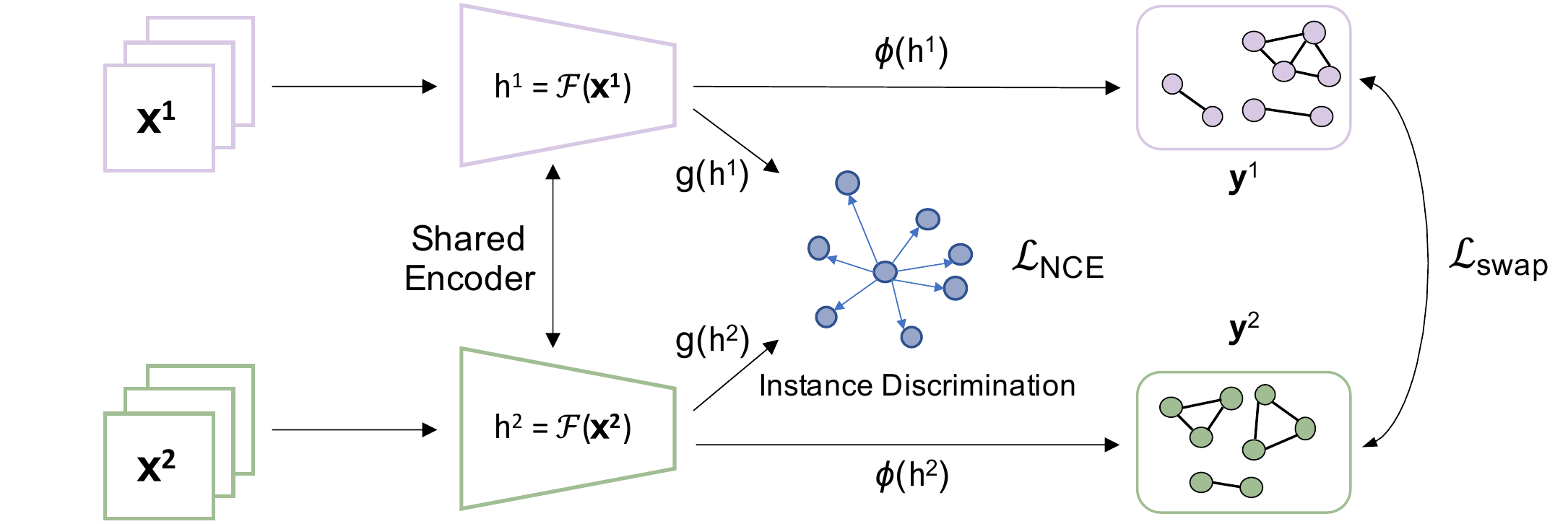}
    \caption{The overall framework of our proposed method. We adopt a two head based structure ($g$ and $\phi$). The first head $g$ will play a regular instance discrimination task. The second head $\phi$ will generate a weak label based on the connected component labeling process, then use the weak label to perform a supervised contrastive learning task. Please see more details in section 3.}
    \label{fig:framework}
    \vspace{-10pt}
\end{figure*}

\section{Related Work}

\textbf{Self-Supervised Learning}. Early work in self-supervised learning mainly focuses on the designing of different pretext tasks. For example, predict a relative offset for a pair of patches \cite{patch_prediction}, solving the jigsaw puzzles \cite{jigsaw}, colorize the gay-scaled images \cite{colorful}, image inpainting \cite{inpainting},  predicting the rotation angle \cite{rotation}, unsupervised deep clustering \cite{deepclustering} and image reconstruction \cite{autoencoder, gan, bigbigan, biggan, srgan}.  Although these methods have shown their effectiveness, they lack the generality of the learned representations.

\textbf{Contrastive Learning}. Contrastive learning \cite{cpc, deepinfomax, instance_discrimination, alignment_uniformity} has become one of the most successful methods in the field of self-supervised learning. As we mentioned, most recent works mainly focus on the augmentation for positive samples and the exploration for negative samples. For example, SimCLR \cite{simclr} proposed composition of data augmentations \eg Grayscale, Random Resized Cropping, Color Jittering, and Gaussian Blur to making the model robust to these transformations.  InfoMin \cite{goodview} further introduced an ``InfoMin principle" which suggests that a good augmentation strategy should reduce the mutual information between the positive pairs while keeping the downstream task-relevant information intact. To explore the use of negative samples, InstDisc \cite{instance_discrimination} proposed a memory bank store the representation of all the images in the dataset. MoCo \cite{moco, mocov2} increasing the number of negatives by using a momentum contrast mechanism that forces the query encoder to learn the representation from a slowly progressing key encoder and maintains a long queue to provide a large number of negative examples.

% \begin{figure*}[t]
%     \centering
%     \includegraphics[width=0.9\linewidth]{Assets/framework.pdf}
%     \caption{The overall framework of our proposed method. We adopt a two head based structure ($g$ and $\phi$). The first head $g$ will play a regular instance discrimination task. The second head $\phi$ will generate a weak label based on the connected component labeling process, then use the weak label to perform a supervised contrastive learning task. Please see more details in section 3.}
%     \label{fig:framework}
% \end{figure*}

\textbf{Contrastive Learning Without Negatives}. Unlike the typical contrastive learning framework, BYOL \cite{byol} can learn a high-quality visual representation without the negative samples. Specifically, it trains an online network to predict the target network representation of the same image under a different augmented view and using an additional predictor network on top of the online encoder to avoiding the model collapse. SimSiam \cite{SimSiam} extends BYOL to explore the siamese structure in contrastive learning further. Surprisingly, SimSiam prevents the model collapse even without the target network and large batch size; although the linear evaluation result is lower than BYOL, it performs better in the downstream tasks.

\section{Method}
In this section, we will first revisit the preliminary work on contrastive learning and address its limitations. Then we will introduce our proposed weakly supervised contrastive learning framework (WCL), which automatically mines similar samples while doing the instance discrimination. After that, the algorithm and the implementation details will also be explained.

\subsection{Revisiting Contrastive Learning}
Typical contrastive learning methods adopt the noise contrastive estimation (NCE) objective for discriminating different instance in the dataset. Concretely, NCE objective encourages different augmentations of the same instance to be pulled closer in a latent space yet pushes away different instances’ augmentations. Following the setup of SimCLR \cite{simclr}, given a batch of unlabeled samples $\{\mathbf{x}\} ^{N}_{i=1}$, we randomly apply a composition of augmentation functions $T(\cdot)$ to obtain two different views of the same instance, which can be written as $\{\mathbf{x}^{1}\}^{N}_{i=1} = T(\mathbf{x}, \theta_{1})$ and $\{\mathbf{x}^{2}\}^{N}_{i=1} = T(\mathbf{x}, \theta_{2})$ where $\theta$ is random seed for $T$. Then, a convolutional neural network based encoder $\mathcal{F}(\cdot)$ will extract the information from different augmentations, that can be expressed by $\{\mathbf{h}^{1}\} = \mathcal{F}(\{\mathbf{x}^{1}\}^{N}_{i=1})$ and $\{\mathbf{h}^{2}\} = \mathcal{F}(\{\mathbf{x}^{2}\}^{N}_{i=1})$. Finally, a non-linear projection head $\mathbf{z} = g(\mathbf{h})$ maps representations $\mathbf{h}$ to the space where the NCE objective is applied.  If we {denote} $(\mathbf{z}_{i}, \mathbf{z}_{j})$ as a positive pair, the NCE objective can be expressed as
\begin{equation}
    \label{equation:nce}
    \mathcal{L}_{NCE} = -\log \frac{\exp(sim(\mathbf{z}_{i}, \mathbf{z}_{j})/ \tau) }{\sum_{k=1}^{N} \mathbb{1}_{[k \neq i]} \exp(sim(\mathbf{z}_{i}, \mathbf{z}_{k}) / \tau ) }.
\end{equation}

\begin{figure*}[t]
    \centering
    \includegraphics[width=0.8\linewidth]{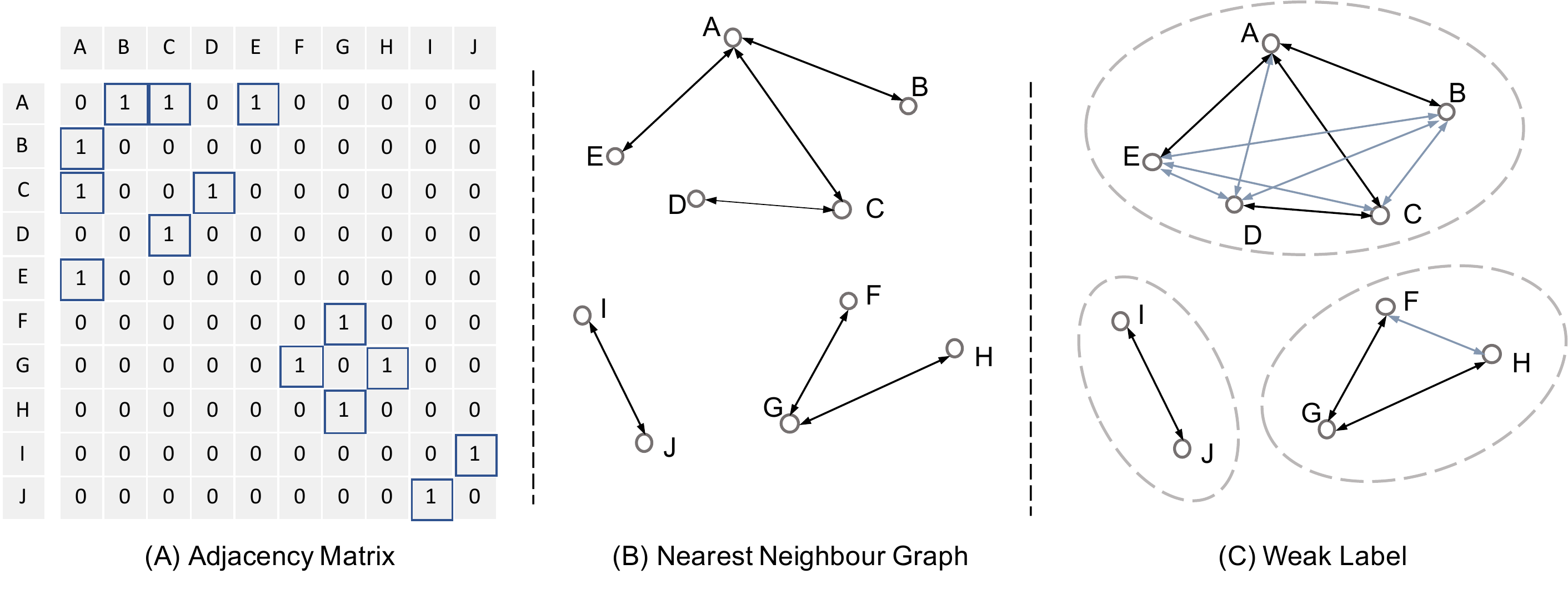}
    \caption{The procedure of weak label generation.}
    \label{fig:graph}
    \vspace{-10pt}
\end{figure*}

\subsection{Instance Similarities as Weak Supervision}
The instance discrimination based methods have already shown promising performance for unsupervised pretraining. However, this line of solution ignores the relationships between different images because only the augmentations from the same image will be regarded as the same class. Inspired by previous works, we can leverage the embedding vectors to explore the relations between different images. Specifically, we will generate a weak label based on the embedding vectors and then use it as a supervisory signal to attract similar samples in the embedding space. However, direct use of weak supervision will cause two problems. First, there is a natural conflict between ``instance discrimination" and ``similar sample attraction" since one wants to push all the different instances away, and the other wants to pull similar samples closer. Second, there might be noise in the weak label, especially in the early training stages. Simply attracting similar samples based on the weak label will slow down the convergence of the model. 

\textbf{Two-head framework.} To resolve these issues, we proposed an auxiliary projection head $\phi(\cdot)$. In this case, the primary projection head $g(\cdot)$ will still perform a regular instance discrimination task to focus on the instance level information; the auxiliary projection head consists of the same structure with $g(\cdot)$ and will explore the similar samples and generate a weak label as the supervisory signal to attract similar samples. With these two heads of distinct responsibilities, we can further transform the features extracted by the encoder $\mathcal{F}$ into different embedding spaces to resolve the conflict. Moreover, the primary projection head will ensure the model's convergence even when the weak label has some noise. The information extracted from the auxiliary projection head can be written as
\begin{equation}
    \label{equation:graph_head}
    \mathbf{v}_{i} = \phi( \mathcal{F} ( T(\mathbf{x}_i, \theta) ) ) ).
\end{equation}
Suppose we have obtained a weak label $\mathbf{y} \in \mathbb{R}^{N \times N}$ based on $\mathbf{v}$ which denotes whether a pair of samples is similar (\ie $\mathbf{y}_{ij} = 1$ means $\mathbf{x}_i$ and $\mathbf{x}_j$ are similar). Different from Eq. (\ref{equation:nce}) that naturally forms positive pairs through augmentations, we can then leverage the label $\mathbf{y}_{ij}$ to indicate whether  $\mathbf{x}_i$ and $\mathbf{x}_j$ can produce a positive pair or not. By introducing an indicator $\mathbb{1}_{\mathbf{y}_{ij}=1}$ into Eq. (\ref{equation:nce}), we achieve the supervised contrastive loss \cite{SupervisedCL}
\begin{align}
    \label{equation:attraction}
    &\mathcal{L}_{sup} = \frac{1}{N}  \sum_{i=0}^{N} \mathcal{L}_{sup}^i \\
    \label{equation:sup}
    \mathcal{L}_{sup}^{i} =  -\sum_{j}^{N} \mathbb{1}_{\mathbf{y}_{ij}=1} & \log \frac{\exp(sim(\mathbf{v}_{i}, \mathbf{v}_{j})/ \tau) }{\sum_{k=1}^{N} \mathbb{1}_{[k \neq i]} \exp(sim(\mathbf{v}_{i}, \mathbf{v}_{k}) / \tau ) },
\end{align}
which has been shown to be more effective than the traditional supervised cross-entropy loss.

\newlength{\textfloatsepsave} 
\setlength{\textfloatsepsave}{\textfloatsep} 
\setlength{\textfloatsep}{5pt}

\subsection{Weak Label Generation}
In this section,  we will elaborate how to generate the weak label for the mini-batch of samples. The overall idea can be summarized into two points: First, for each sample, the closest sample can be regarded as a similar sample. Second, if $(\mathbf{x}_i,  \mathbf{x}_j)$ and $(\mathbf{x}_j,  \mathbf{x}_k)$  are two pairs of similar samples, then we can think that $\mathbf{x}_i$ and $\mathbf{x}_k$ are also similar.

Suppose we use the auxiliary projection head $\phi$ to map a batch of samples to $N$ embeddings $V = \{\mathbf{v}_{1}, \mathbf{v}_{2}, ...,  \mathbf{v}_{N} \}$.  Then, for each sample $\mathbf{v}_i$, we find the closest sample $\mathbf{v}_j$ by computing the cosine similarity score. Now, we can define an adjacency matrix by:
\begin{equation}
    \label{equation:adjacency}
    A(i, j) = 
    \begin{cases}
        1, & \text{if $i = k^{1}_{j}$ or $j = k^{1}_{i}$} \\
        0, & \text{otherwise} 
    \end{cases}
\end{equation}
Here, we use $k^{1}_{i}$ to denote the 1-nearest neighbour of $\mathbf{v}_i$. Basically, Eq.\eqref{equation:adjacency} will generate a sparse and symmetric 1-nearest neighbor graph where each vertex is linked with its closest sample. To find out all similar samples, we can convert this problem into a Connected Components Labeling (CCL) process; that is, for each sample, we want to find all the reachable samples based on the 1-nearest neighbor graph. This is a traditional graph problem that can be easily solved by the famous Hoshen–Kopelman algorithm \cite{twopass} (also known as the two-pass algorithm). We define an undirected graph by $G = (V, E)$ where $V$ is the embedding from $\phi$, and edges $E$ connecting the vertex $A(i, j) = 1$. The algorithm adopts a Disjoint-set data structure that consists of three operations: \textbf{makeSet}, \textbf{union} and \textbf{find}  (see the definition in Algorithm \ref{alg:ccl}). Basically, it first creates a singleton set for each $\mathbf{v}$ in $V$, then traverses each edge in $E$ and merges different sets through the edges; finally, it returns the set for each vertex that belongs to.  Back to our proposed idea, we will treat the samples in the same set as similar samples. Now, the weak label can be defined as:
\begin{equation}
    \label{equation:weaklabel}
    \mathbf{y}_{ij} = 
    \begin{cases}
        1, & \text{if find($\mathbf{v}_i$) = find($\mathbf{v}_j$) and $i \neq j$ } \\
        0, & \text{otherwise} 
    \end{cases}
\end{equation}
Such weak label generation method has several advantages.
\begin{itemize}
 \vspace*{-0.3em}
 \setlength\itemsep{-0.1em}
 \item This is a parameter-free process, so we do not need any hyperparameter optimization.
 
 \item Based on the definition of an undirected graph and connected components, the weak label is always reciprocal. (\ie $\mathbf{y}_{ij} = \mathbf{y}_{ji}$)
 
 \item This is a deterministic process; the final result does not depend on any initial state.
\end{itemize}
The weak label will be used as the supervisory signal for the auxiliary projection head $\phi$. However, if $\mathbf{v}_i$ and $\mathbf{v}_j$ are in the same set,  $sim(\mathbf{v}_i$, $\mathbf{v}_j)$ is very likely to be a large number. According to Eq. \eqref{equation:sup}, directly using the weak label will cause $\mathcal{L}_{sup}$ to be very small, which is not conducive to the model's optimization. To resolve this issue, we can simply swap the weak label to supervise the same batch of samples with different augmentations.  Concretely, we derive embeddings $V^{1}$ and $V^{2}$ from two types of augmentations, based on which we generate the corresponding weak label $\mathbf{y}^{1}$, $\mathbf{y}^{2}$. Then $\mathbf{y}^{1}$ will be used as the supervisory signal for $V^{2}$ and vice versa. The swapped version of Eq. \eqref{equation:attraction} can be written as:
\begin{equation}
    \label{equation:swap_ce}
    \mathcal{L}_{swap} = \mathcal{L}_{sup}(V^1, \mathbf{y}^2) +  \mathcal{L}_{sup}(V^2, \mathbf{y}^1).
\end{equation}

\begin{algorithm}[t]
\SetAlgoLined
\SetKwInOut{Input}{Input}
\Input{An adjacency matrix $G = (V, E)$}

Define makeSet(v) : Create a new set with element v

Define union(A, B): Return the set $A \cup B$

Define find(v): Return the set which contains v

\For{v in V}{
    makeSet(v)
}
\For{each ($v_i$, $v_j$) in E}{
    \If{ find($v_i$) $\neq$ find($v_j$)}{ 
        union(find($v_i$), find($v_j$)) 
    }
}
\For{each v in V} {
    return the set contains v:  find(v)
}
\SetKwInOut{Output}{Output}
\Output{The corresponding identification of the connected component for each $\mathbf{v}$.}
\caption{Connected Components Labeling}
\label{alg:ccl}
\end{algorithm}

\subsection{Label Propagation with Multi-Crops}
Since the comparison between random crops of an image plays the key role in contrastive learning, there are lots of previous works \cite{DCL} pointing out that increasing the number of crops or views can significantly increase the representation quality. SwAV \cite{swav} introduced a multi-crop strategy that adds K additional low-resolution crops in each batch. Using low-resolution images can greatly reduce computational costs. However, the multiple crops of the same image may have many overlap areas. In this case, more crops may not provide additional effective information. To address this issue, we proposed a K-Nearest Neighbor based Multi-crops strategy. Specifically, we will store the feature $\mathbf{h}^{1}$ for every batch and then use these features to find the $K$ closest samples based on the cosine similarity at the end of each epoch. Finally, we will use the low-resolution crops of the K closest images in the next epoch. If we apply the $\mathcal{L}_{swap}$ on the K-NN multi-crops, the number of positive samples can be expended to K times. Note that the K-NN result is unreliable in the early training; hence, we should use the standard multi-crops strategy to warm up the model for a certain number of epochs and then switch to our K-NN multi-crops to get better performance. (See more details in our experiments.) If we use $\mathcal{L}_{cNCE}$ and $\mathcal{L}_{cswap}$ to denote the contrastive loss and swapped loss for the multi-crops images, then the overall training objective for our weakly supervised contrastive learning framework can be expressed as
{\small
\begin{align}
    \begin{split}
    \label{equation:overall}
    \mathcal{L}_{overall} = \mathcal{L}_{NCE} + \lambda\mathcal{L}_{cNCE} +  \beta\mathcal{L}_{swap}  + \gamma\mathcal{L}_{cswap},
    \end{split}
\end{align}
}\noindent
where $\lambda$, $\beta$ and $\gamma$ are the hyper-parameters. We simply take $\lambda=1$, $\beta=0.5$ and $\gamma=0.5$ in our implementation. Please see more details in Algorithm \ref{alg:overall}.

\begin{algorithm}[t]
\SetAlgoLined
\SetKwInOut{Input}{Input}
\Input{$\{\mathbf{x}^1\}^{N}_{i=1}$ and $\{\mathbf{x}^2\}^{N}_{i=1}$: a batch of samples with different augmentations. $\mathcal{F}$: the backbone network. $g$: the first projection head. $\phi$: the auxiliary projection head.}

\While{network not converge} {
    Initialize an empty list L ;
    
    \For{i=1 to step}{
        $\mathbf{h}^1 = \mathcal{F}(\{\mathbf{x}^1\}^{N}_{i=1})$ \quad
        $\mathbf{h}^2 = \mathcal{F}(\{\mathbf{x}^2\}^{N}_{i=1})$
        
        $\mathbf{z}^1 = g(\mathbf{h}^1)$ $\quad\quad\quad\;\;$ $\mathbf{z}^2 = g(\mathbf{h}^2)$
        
        $\mathbf{v}^1 = \phi(\mathbf{h}^1)$ $\quad\quad\quad\;$ $\mathbf{v}^2 = \phi(\mathbf{h}^2)$
        
        Calculate contrastive loss $\mathcal{L}_{NCE}$  Eq. \eqref{equation:nce}
        
        Generate weak label $\mathbf{y}^1$, $\mathbf{y}^2$ based on $\mathbf{v}^1$, $\mathbf{v}^2$
        
        Calculate swapped loss $\mathcal{L}_{swap}$  Eq. \eqref{equation:swap_ce}
        
        Calculate $\mathcal{L}_{cNCE}$ and $\mathcal{L}_{cswap}$ 
        
        Optimize the network by $\mathcal{L}_{overall}$ Eq. \eqref{equation:overall}
        
        Append $\mathbf{h}^1$ to list L ;
    }
    
    Compute the $K$-NN for each sample based on L.
}

\SetKwInOut{Output}{Output}
\Output{The well trained model $\mathcal{F}$}
\caption{Weakly Supervised Contrastive Learning (WCL)}
\label{alg:overall}
\end{algorithm}

\begin{table*}[t]
 \centering
 \small
 \caption{Experiments on CIFAR-10 and CIFAR-100 with different batch size and training epochs}
 \label{table:small_dataset}
 \begin{tabular}{ |c || c || c | c | c | c | c | c | c | c | } 
 \hline
 \multirow{2}{*}{Batch Size} & \multirow{2}{*}{ Method } & 
 \multicolumn{4}{c|}{CIFAR10} & \multicolumn{4}{c|}{CIFAR100} \\
  \cline{3-10} & & 100 ep & 200 ep & 300 ep & 400 ep & 100 ep & 200 ep & 300 ep & 400 ep \\

 \hline \hline
 64   &  SimCLR  & 77.20  & 80.64 & 82.77 & 84.48 & 52.35  & 55.86 & 58.18 & 59.96 \\ \hline
 64   &  WCL (Ours)    & \makecell{79.17 \\ \textbf{\textcolor{improvement}{(+1.97)}}}   & \makecell{83.54 \\ \textbf{\textcolor{improvement}{(+2.90)}}} & \makecell{85.68 \\ \textbf{\textcolor{improvement}{(+2.91)}}} & \makecell{86.64 \\ \textbf{\textcolor{improvement}{(+2.16)}}} & \makecell{53.54 \\ \textbf{\textcolor{improvement}{(+1.19)}}}  & \makecell{56.57 \\ \textbf{\textcolor{improvement}{(+0.71)}}} & \makecell{59.29 \\ \textbf{\textcolor{improvement}{(+1.11)}}} & \makecell{60.76 \\ \textbf{\textcolor{improvement}{(+0.80)}}} \\ \hline \hline
 128  &  SimCLR  & 79.64  & 83.57 & 85.70 & 86.72 & 54.72  & 59.19 & 60.88 & 62.20 \\ \hline
 128   &  WCL (Ours)    & \makecell{81.82 \\ \textbf{\textcolor{improvement}{(+2.18)}}}   & \makecell{85.65 \\ \textbf{\textcolor{improvement}{(+2.08)}}} & \makecell{87.81 \\ \textbf{\textcolor{improvement}{(+2.91)}}} & \makecell{88.65 \\ \textbf{\textcolor{improvement}{(+1.93)}}} & \makecell{55.46 \\ \textbf{\textcolor{improvement}{(+0.74)}}}  & \makecell{60.30 \\ \textbf{\textcolor{improvement}{(+1.11)}}} & \makecell{61.73 \\ \textbf{\textcolor{improvement}{(+0.85)}}} & \makecell{63.17 \\ \textbf{\textcolor{improvement}{(+0.97)}}} \\ \hline \hline
 256  &  SimCLR  & 81.78  & 85.34 & 87.29 & 88.48 & 57.16  & 61.18 & 63.49 & 64.20 \\ \hline
 256   &  WCL (Ours)    & \makecell{83.12 \\ \textbf{\textcolor{improvement}{(+1.34)}}}   & \makecell{87.57 \\ \textbf{\textcolor{improvement}{(+2.23)}}} & \makecell{88.85 \\ \textbf{\textcolor{improvement}{(+1.56)}}} & \makecell{89.47 \\ \textbf{\textcolor{improvement}{(+0.98)}}} & \makecell{57.85 \\ \textbf{\textcolor{improvement}{(+0.70)}}}  & \makecell{62.98 \\ \textbf{\textcolor{improvement}{(+1.80)}}} & \makecell{64.21 \\ \textbf{\textcolor{improvement}{(+0.72)}}} & \makecell{64.93 \\ \textbf{\textcolor{improvement}{(+0.73)}}} \\ \hline
 \end{tabular}
 \vspace{-10pt}
\end{table*}

\section{Experimental Results}
\subsection{Ablation Studies}

In this section, we will empirically study our Weak Supervised Contrastive Learning (WCL) framework under different batch sizes, epochs, datasets(CIFAR-10, CIFAR-100, ImageNet100) and show the effectiveness of each component by extensive experiments. 

\textbf{CIFAR-10 and CIFAR-100.} The CIFAR-10 \cite{cifar} dataset consists of 60000 32x32 colour images in 10 classes, with 6000 images per class. There are 50000 training images and 10000 test images. CIFAR-100 is just like the CIFAR-10, except it has 100 classes containing 600 images each. There are 500 training images and 100 testing images per class. We use the ResNet50 \cite{resnet} as our backbone network. Because the training images only contain 32x32 pixels, we replace the first 7x7 Conv of stride 2 with 3x3 Conv of stride 1 and also remove the first max pooling operation. We use 2-Layer-MLP for the two non-linear projection heads. For data augmentation, we use the random resized crops (the lower bound of random crop ratio is set to 0.2), color distortion (strength=0.5), and leaving out Gaussian blur. The model is trained using LARS optimizer \cite{lars} with a momentum of 0.9 and weight decay of $1e^{-6}$. We linear warm up the learning rate for 10 epochs until it reaches $0.25 \times BatchSize / 256$, then switch to the cosine decay scheduler \cite{cosine_lr}. The temperature parameter $\tau$ is always set to $0.1$. To perform the Connected Components Labeling process, we simply use the ``connected\_components" function from the Scipy Library \cite{scipy}. We will use the same training strategy for both CIFAR-10 and CIFAR-100.

\setlength{\textfloatsep}{\textfloatsepsave}

\textbf{ImageNet-100.} ImageNet-100 dataset is a randomly chosen subset from ILSVRC2010 ImageNet \cite{imagenet_cvpr09}. (We simply take the first 100 class in our experiments.) For training the ImageNet-100, we strictly follow the training strategy reported in SimCLR \cite{simclr}. Specifically, we set the $BatchSize=2048$, and use the LARS optimizer with $lr=0.075\times \sqrt{BatchSize}$. Moreover, we found that the default augmentation that used in SimCLR might be too strong, which makes the model very hard to converge in the beginning; thus, we adopt the same but a little bit weaker version of the augmentation (the one that used in MoCoV2\cite{mocov2}) in the first 10 epochs and then switch it back to the original augmentations after warm-up. The model will be optimized for 200 epochs, and the rest of the settings (including temperature, weight decay, etc.) are the same as our CIFAR training.

\textbf{Evaluation Protocol}. For testing the representation quality, we evaluate our well-trained model on the widely adopted linear evaluation protocol - We will freeze the encoder parameters and train a linear classifier on top of it by using the standard SGD optimizer with a momentum of 0.9, learning rate of $0.1 \times BatchSize / 256$ and cosine decay scheduler. We don't use any regularization techniques such as weight decay and gradient clipping. The model will be trained for 80 epochs, then evaluated on the testing set.

\textbf{Effect of weak supervision}. We choose SimCLR as our baseline, and compare it with our method on $BatchSize=64, 128, 256$ and $Epoch=100, 200, 300, 400$. Note, in these experiments; we do not use any multi-crops strategy; only an additional $\mathcal{L}_{swap}$ is applied on top of the SimCLR. Table \ref{table:small_dataset} shows the results. Obviously, our proposed method substantially outperforms the baseline across all settings. For CIFAR-10, we have various improvements from $0.98\%$ to $2.91\%$ based on different settings. For CIFAR-100, the improvement is from $0.73\%$ to $1.80\%$.

\begin{table}[h]
\small
 \caption{Effectiveness of two-head framework (ImageNet100)}
 \label{table:ablation}
\begin{tabular}{ |c | c | c | c | c | c | c |} 
\hline
$g$ & $\phi$ & $\mathcal{L}_{NCE}$ & $\mathcal{L}_{swap}$ & $\mathcal{L}_{cNCE}$ & $\mathcal{L}_{cswap}$ & Top-1  \\
\hline
\checkmark & & \checkmark & & &  & 75.79 \\
 & \checkmark & & \checkmark &  &  & 71.33 \\
\checkmark & & \checkmark & \checkmark & &  & 75.26 \\
\checkmark & \checkmark & \checkmark & \checkmark & &  & \textbf{77.51} \\ \hline
\checkmark & \checkmark & \checkmark & \checkmark & \checkmark &  & 79.06 \\
\checkmark & \checkmark & \checkmark & \checkmark &  & \checkmark & 79.08 \\
\checkmark & \checkmark & \checkmark & \checkmark & \checkmark & \checkmark & \textbf{79.77} \\ \hline
\end{tabular}
\end{table}

\textbf{Effect of two-head framework}. We also perform an extensive ablation study to examine the effectiveness of our two head based framework. The experiments are mainly performed on the ImageNet-100 dataset, and the result is shown in Table \ref{table:ablation}. Note, the $\mathcal{L}_{cNCE}$ and $\mathcal{L}_{cswap}$ in this experiment is based on the standard multi-crops strategy (without KNN). The first row is the SimCLR baseline. The second row is the case that only $\mathcal{L}_{swap}$ is applied; the model can still learn a meaningful representation but result in a worse accuracy than the baseline. We also try to apply both $\mathcal{L}_{NCE}$ and $\mathcal{L}_{swap}$ on the same head; from the third row, we can see there is a $0.53\%$ performance drop. We doubt this is because of the conflicts between the instance discrimination and similar sample attraction. The fourth row shows our proposed method, which separates the two tasks on different heads. In this case, we get $1.72\%$ improvements over the baseline, which verified our hypothesis. The last three rows show the result with the multi-crops strategy, and the performance can be further improved by $2.26\%$.

\textbf{Effect of K-NN Multi-Crops}. As we have mentioned, the K-NN result is unreliable in the early training, and we need to use the standard multi-crops strategy to warm up the model for a certain number of epochs. Table \ref{table:pretrain} shows the result for a different number of warm up epochs. We can see clearly that with 50 epochs of warm up, our K-NN multi-crops strategy has $1\%$ improvements over the standard multi-crops (see the last row in Table \ref{table:ablation}). Finally, our proposed method achieved $80.78\%$ Top-1 accuracy on linear evaluation, which has $5\%$ improvements than the SimCLR baseline ($75.79\%$).

\begin{table}[h]
 \vspace{-3pt}
 \centering
 \small
 \caption{Warm up epochs for K-NN Multi-Crops (K=4)}
 \label{table:pretrain}
\begin{tabular}{c | c | c | c | c | c  } 
\hline
Epochs & 0 & 25 & 50 & 75 & 100 \\
\hline
Accuracy &  79.73 & 80.25 & \textbf{80.78} & 80.63 & 80.23 \\
\hline
\end{tabular}
\vspace{-5pt}
\end{table}

\begin{figure}[h]
    \vspace{-4mm}
    \centering
    \includegraphics[width=0.9\linewidth]{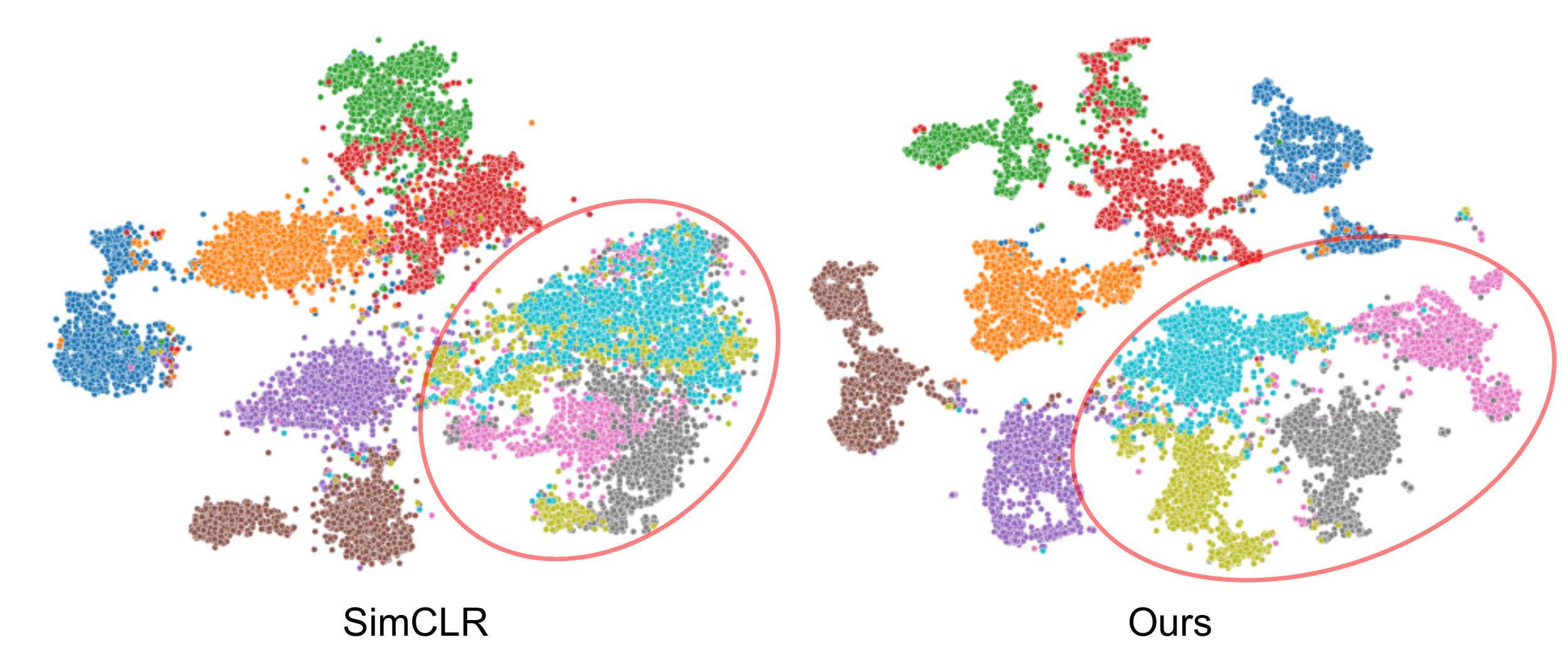}
    \caption{t-SNE visualization for SimCLR and our method}
    \label{fig:tsne}
    \vspace{-4mm}
\end{figure}

\textbf{Visualization}. Figure \ref{fig:tsne} shows the t-SNE visualization \cite{tsne} of $\mathbf{h}$ from a randomly selected 10 classes. Compare to SimCLR; our weakly supervised contrastive learning framework can enhance a much better intra-class compactness and inter-class discrepancy.

\subsection{Comparison on ImageNet-1K Dataset}
We also performed our algorithm on the large-scale ImageNet-1k dataset \cite{imagenet_cvpr09}. The training strategy is the same as our ImageNet-100 training, except we adopt a larger batch size (4096) and use the 3-Layer-MLP for the two projection heads. For the K-NN Multi-crops, we simply take the best strategy from Table \ref{table:pretrain}, which means we will use the standard multi-crops strategy for the first 25\% epochs, and then switch to our K-NN version.

\begin{table}[h]
 \centering
 \small
 \caption{Compare to FNCancel on ImageNet-1K}
 \label{table:compare_fncancel}
\begin{tabular}{l  c  c  c} 
\toprule 
Method & Epochs & GPU(time) & Acc \\
\hline
SimCLR & 100 & 1.00 & 66.4 \\ 
FNCancel & 100 & - & 68.1 \\
WCL (Ours) & 100 & 1.01 & 68.1 \\
SimCLR & 1000 & 10.00 & 70.3 \\ 
FNCancel + multi-crops & 100 & 2.85 & 70.4 \\
WCL (Ours) + multi-crops & 100 & 1.31 & \textbf{71.0} \\
\toprule 
\end{tabular}
\vspace{-3mm}
\end{table}

\textbf{Compare to FNCancel}. \cite{FNCancel} Table \ref{table:compare_fncancel} shows the comparison between our proposed method with FNCancel and SimCLR.  Note, for a fair comparison, all models are trained with a 3-Layer-MLP projection head. As we can see, with a negligible additional computational cost (0.01), our proposed method can surpass the SimCLR baseline $1.7\%$ and achieved the same result with FNCancel. FNCancel does not report the standard time usage on the paper, but since it requires 8 additional forward passes to generate the support view embeddings, their actual computational cost will be much higher than ours. We also compare the result with the multi-crops strategy. In this case, we use 2 $160\times160$ images as our main views and 6 additional $96\times96$ K-NN crops. Look at the last row; our proposed method can achieve 71.0 top-1 accuracy with only $31\%$ more additional cost than SimCLR. This is twice faster than FNCancel and has 0.6\% improvements on linear evaluation.

\begin{table}[h]
 \centering
 \small
 \vspace{-1mm}
 \caption{Top-1 accuracy under the linear evaluation on ImageNet with the ResNet-50 backbone. The table compares the methods over 200 epochs of pretraining. * denotes multi-crops strategy.}
 \label{table:200epoch}
\begin{tabular}{l  c  c c c} 
\toprule 
Method & Arch & Param & Epochs & Top-1 \\
Supervised & R50 & 24 & - & 76.5 \\ \hline
InstDisc \cite{instance_discrimination} & R50 & 24 & 200 & 58.5 \\
LocalAgg \cite{local} & R50 & 24 & 200 & 58.8 \\
SimCLR \cite{simclr} & R50 & 24 & 200 & 66.8 \\
MoCo \cite{moco} & R50 & 24 & 200 & 60.8 \\
MoCo v2 \cite{mocov2} & R50 & 24 & 200 & 67.5 \\
MoCHi \cite{mochi} & R50 & 24 & 200 & 68.0 \\
CPC v2 \cite{cpc} & R50 & 24 & 200 & 63.8 \\
PCL v2 \cite{PCL} & R50 & 24 & 200 & 67.6 \\
SimSiam \cite{SimSiam} & R50 & 24 & 200 & 70.0 \\
SwAV \cite{swav} & R50 & 24 & 200 & 69.1 \\
SwAV* \cite{swav}  & R50 & 24 & 200 & 72.7 \\ \hline
WCL (Ours) & R50 & 24 & 200 & 70.3 \\
WCL* (Ours) & R50 & 24 & 200 & \textbf{73.3} \\ \toprule 
\end{tabular}
\vspace{-15pt}
\end{table}

\begin{table}[h]
 \centering
 \small
 \caption{Top-1 accuracy under the linear evaluation on ImageNet. The table compares the methods with more epochs of pretraining. * denotes multi-crops strategy.}
 \label{table:800epoch}
\begin{tabular}{l  c  c  c  c} 
\toprule 
Method & Arch & Param & Epochs & Top-1 \\
Supervised & R50 & 24 & - & 76.5 \\ 
\hline
SeLa \cite{Self-labelling} & R50 & 24 & 400 & 61.5 \\
SimCLR \cite{simclr} & R50 & 24 & 800 & 69.1 \\
SimCLR v2 \cite{simclrv2} & R50 & 24 & 800 & 71.7 \\
MoCo v2 \cite{mocov2} & R50 & 24 & 800 & 71.1 \\
SimSiam \cite{SimSiam} & R50 & 24 & 800 & 71.3 \\
SwAV \cite{swav} & R50 & 24 & 800 & 71.8 \\
BYOL \cite{byol} & R50 & 24 & 1000 & 74.3 \\
FNCancel* \cite{FNCancel} & R50 & 24 & 1000 & 74.4 \\
AdpCLR \cite{topk} & R50 & 24 & 1100 & 72.3 \\ 
\hline
WCL (Ours) & R50 & 24 & 800 & 72.2 \\
WCL* (Ours) & R50 & 24 & 800 & \textbf{74.7} \\ 
\hdashline
\emph{Others} \\
SwAV* \cite{swav} & R50 & 24 & 800 & \textbf{75.3} \\
\toprule 
\end{tabular}
\vspace{-5pt}
\end{table}

\begin{table*}[t]
 \vspace{-5mm}
 \centering
 \small
 \caption{Low-shot image classification on VOC07}
 \label{table:low-shot}
\begin{tabular}{l  c c  c  c  c  c  c  c  c  c} 
\toprule 
Method & Epochs & k=1 & k=2 & k=4 & k=8 & k=16 & k=32 & k=64 & Full \\
Random & - & 8.92  & 9.33 & 10.10  & 10.42 & 10.82 & 11.34 & 11.96 & 12.42  \\
Supervised & 90 & 54.46  & 68.15 & 73.79  & 79.51 & 82.26 & 84.00 & 85.13 & 87.27  \\
\hline 
MoCo v2 \cite{mocov2} & 200 &  46.30 & 58.40  & 64.85 & 72.47  & 76.14 & 79.16 & 81.52 & 84.60   \\
PCL v2 \cite{PCL} & 200 &  47.88 & 59.59  & 66.21 & 74.45  & 78.34 & 80.72 & 82.67 & 85.43 \\
SwAV \cite{swav} & 200 &  43.07 & 55.65  & 64.82 & 73.17  & 78.38 & 81.86 & 84.40 & 87.47 \\
WCL (Ours) & 200 & \textbf{48.06}  & \textbf{60.12} & \textbf{68.52}  & \textbf{76.16}  & \textbf{80.24} & \textbf{82.97} &  \textbf{85.01}  & \textbf{87.75}  \\ 
\hline 
SwAV \cite{swav} & 400 &  42.14 & 55.34  & 64.31 & 73.08  & 78.47 & 82.09 & 84.62 & 87.78 \\
SwAV \cite{swav} & 800 &  42.85 & 54.90  & 64.03 & 72.94  & 78.65 & 82.32 & 84.90 & 88.13 \\
WCL (Ours) & 800 & \textbf{48.25}  & \textbf{60.68} & \textbf{68.52}  & \textbf{76.48}  & \textbf{81.05} & \textbf{83.89} &  \textbf{85.88}  & \textbf{88.64}  \\ 
\toprule 
\end{tabular}
\vspace{-5mm}
\end{table*}

\textbf{Linear Evaluation}. 
For the linear evaluation of ImageNet-1k, we strictly follow the setting in SimCLR \cite{simclr}. Table \ref{table:200epoch} and \ref{table:800epoch} shows our result for 200 epochs and 800 epochs of training. We also report the result with 2 $224\times224$ and 6 additional $96\times96$ K-NN crops (as in SwAV \cite{swav}). We can see clearly that when the model is optimized for 200 epochs, our proposed method achieved state-of-the-art performance among all the recent self-supervised learning frameworks. When the model is trained for 800 epochs, our model can still outperform most recent works but slightly lower than SwAV.

\begin{table}[h]
 \centering
 \small
 \caption{ImageNet semi-supervised evaluation.}
 \label{table:semi}
\begin{tabular}{l  c  c  c  c} 
\toprule 
 & 1\% &  & 10\% &  \\
Method & Top-1 & Top-5 & Top-1 & Top-5 \\
Supervised & 25.4 & 56.4 & 48.4 & 80.4 \\ 
\hline
\emph{Semi-supervised} \\
\hline
S4L \cite{s4l} & - & 53.4 & - & 83.8 \\
UDA \cite{uda} & - & 68.8 & - & 88.5 \\
FixMatch \cite{fixmatch} & - & - & 71.46 & 89.1 \\
\hline
\emph{Self-Supervised} \\
\hline
\emph{From AvgPool} \\
InstDisc \cite{instance_discrimination} & - & 39.2 & - & 77.4 \\
PCL \cite{PCL} & - & 75.6 & - & 86.2 \\
PIRL \cite{pirl} & 30.7 & 60.4 & 57.2 & 83.8 \\
SimCLR v1 \cite{simclr} & 48.3 & 75.5 & 65.6 & 87.8 \\ 
BYOL \cite{byol} & 53.2 & 78.4 & 68.8 & 89.0 \\ 
SwAV \cite{swav} & 53.9 & 78.5 & 70.2 & 89.9 \\ 
WCL (Ours) & \textbf{58.3} & \textbf{79.9} & \textbf{71.1} & \textbf{90.3} \\ 
\hdashline
\emph{From Projection Head} \\
SimCLR v2 (R50) \cite{simclrv2} & 57.9 & - & 68.4 & - \\ 
SimCLR v2 (R101)\cite{simclrv2} & 62.1 & - & 71.4 & - \\ 
FNCancel \cite{FNCancel} & 63.7 & 85.3 & 71.1 & 90.2 \\ 
WCL (Ours) & \textbf{65.0}  & \textbf{86.3} & \textbf{72.0} & \textbf{91.2} \\ 
\toprule 
\end{tabular}
\vspace{-10pt}
\end{table}

\textbf{Semi-Supervised Learning}. 
Next, we evaluate the performance obtained when fine-tuning the model representation using a small subset of labeled data. For a fair comparison, we take the same labeled list from SimCLR \cite{simclr}. Specifically, we report our results on two different settings. First, we follow the strategy in PCL \cite{PCL}, and fine-tuning from the average pooling layer of the ResNet50 \cite{resnet} network. In this setting, our model outperforms the previous state-of-the-art (SwAV) 4.4\% on 1\% labels and 0.9\% on 10\% labels.  Then, we also follow the strategy in SimCLRv2 \cite{simclrv2} to fine-tuning from the first layer of the projection head. In this case, our method has 1.3\% and 0.9\% improvement on 1\% and 10\% labels over FNCancel. Notably, this result is even higher than the SimCLRv2 with ResNet101 backbone.

\textbf{Transfer Learning}. 
Finally, We further evaluate the quality of the learned representations by transferring them to other datasets. Following \cite{PCL, swav}, we perform linear classification on the PASCAL VOC2007 dataset \cite{pascal-voc-2007}. Specifically, we resize all images to 256 pixels along the shorter side and taking a 224 × 224 center crop. Then, we train a linear SVM on top of corresponding global average pooled final representations. To study the transferability of the representations in few-shot scenarios, we vary the number of labeled examples k and report the mAP. Table \ref{table:low-shot} shows the comparison between our method with previous works. We report the average performance over 5 runs (except for k=full). The result of our method and SwAV are both based on the multi-crop version. When the model has 200 epochs of pretraining, our method and SwAV can already outperform the supervised pretraining on the full dataset. Interestingly, our method is significantly better than all other works, especially when k is small. When the model has more pretraining epochs, our method can even surpass the supervised pretraining with k = 64 and consistently has higher performance than SwAV across all different k values.
\section{Conclusion}
In this work, we proposed a weakly supervised contrastive learning framework that consist of two projection heads, one of which focus on the instance discrimination task, and the other head adopts the Connected Components Labeling process to generate a weak label, then perform the supervised contrastive learning task by swapping the weak label to different augmentations. Finally, we introduced a new K-NN based multi-crops strategy which has much more effective information and expanding the number of positive samples to K times. Experiments on CIFAR-10, CIFAR-100, ImageNet-100 show the effectiveness of each component. The results of semi-supervised learning and transfer learning demonstrate the state-of-the-art performance for unsupervised representation learning.

\section*{Acknowledgment}
This work is funded by the National Key Research and Development Program of China (No. 2018AAA0100701) and the NSFC 61876095. Chang Xu was supported in part by the Australian Research Council under Projects DE180101438 and DP210101859. Shan You is supported by Beijing Postdoctoral Research Foundation.
%\newpage
{\small
\bibliographystyle{ieee_fullname}
\bibliography{egbib}
}

% \newpage
% \onecolumn
% \maketitle
% \appendix
% \input{Section/supplementary}

\end{document}